\documentclass[letter, 10pt, conference]{ieeeconf} 

\makeatletter
\let\NAT@parse\undefined
\makeatother

\IEEEoverridecommandlockouts                              

\usepackage{times}
\usepackage[numbers]{natbib}
\usepackage{multicol}
\usepackage[bookmarks=true]{hyperref}
\usepackage{color}
\usepackage{hyperref}
\usepackage{textcomp}
\usepackage{amsmath,amsfonts,amssymb}
\usepackage{graphicx}
\usepackage{makecell}
\usepackage{tabularx}
\usepackage{graphicx}
\usepackage{pgfplots}
\usepackage{tikz}
\usepackage{array}
\usepackage{subcaption}
\usepackage{cleveref}
\usepackage{siunitx}
\usepackage[boxed,linesnumbered,vlined]{algorithm2e}
\newcolumntype{?}{!{\vrule width 1pt}}

\definecolor{bblue}{HTML}{4F81BD}
\definecolor{rred}{HTML}{C0504D}

\renewcommand{\baselinestretch}{0.995}

\hypersetup{
  colorlinks   = true, 
  urlcolor     = blue, 
  linkcolor    = black, 
  citecolor   = black 
}

\tikzset{level 1/.style={level distance=0.7cm, sibling distance=1.0cm}}
\tikzset{level 2/.style={level distance=0.7cm, sibling distance=4.0cm}}

\tikzset{bag/.style={draw, rectangle, solid, text width=7em, text centered, yshift=-0.2cm}}

\newcommand{\suav}[1]{sUAV#1\xspace}

\title{ArduSoar: an Open-Source Thermalling Controller\\ for Resource-Constrained Autopilots}

\author{\authorblockN{Samuel Tabor}
\authorblockA{Glasgow, Scotland\\samuelctabor@gmail.com
}
\and
\authorblockN{Iain Guilliard}
\authorblockA{Australian National University\\ Canberra, Australia\\iain.guilliard@anu.edu.au
}
\and
\authorblockN{Andrey Kolobov}
\authorblockA{Microsoft Research\\ Redmond, WA-98052\\akolobov@microsoft.com}
}

\begin{document}
 
\maketitle

\begin{abstract}
Autonomous soaring capability has the potential to significantly increase time aloft for fixed-wing UAVs. In this paper, we introduce ArduSoar, the first soaring controller integrated into a major autopilot software suite for small UAVs. We describe ArduSoar from the algorithmic standpoint, outline its integration with the ArduPlane autopilot, discuss parameter tuning for it, and conduct a series of flight tests on real sUAVs that show ArduSoar's robustness even in highly non-ideal atmospheric conditions. 
\end{abstract}
\section{Introduction}

Autonomous soaring has long been considered for its promise of extending the range and time in the air for Uninhabited Aerial Vehicles (UAVs). These two UAV characteristics are crucial in a range of scenarios from aerial mapping to crop monitoring to package delivery, and are currently constrained by the UAV's powerplant and the limited energy available onboard. Exploiting the energy of the moving air masses has the potential to increase fixed-wing drone range by many times \cite{allen-aiaa06}. While there are many atmospheric phenomena that can help an aircraft gain altitude without the help of a motor, the most ubiquitous of them are \emph{thermals}, rising plumes of air generated by certain areas on the ground giving up heat.

Including the first feasibility study of autonomous thermalling in 1998 \cite{wharington-1998}, researchers have proposed at least 5 approaches to modeling thermals computationally and soaring in them, evaluating these techniques in simulation \cite{wharington-1998,hazard-aiaa10,reddy-pnas16} and in live tests \cite{allen-2007,edwards-aiaa2008,andersson-jgd12}. Each has a significant number of intricacies related to implementation and parameter tuning. Since none of the implementations, written in different languages and for different platforms and setups, are publicly available, none of these techniques have been systematically compared against each other.

This paper introduces a soaring controller called \emph{ArduSoar} and describes its implementation as part of a popular open-source small-UAV (\suav{}) autopilot system, ArduPilot \cite{ardupilot}. ArduPilot can be built for many popular autopilot hardware platforms and run even on the most resource-constrained of them, including Pixhawk \cite{pixhawk}, APM, and a software-in-the-loop simulation. Being part of an open-source project, ArduSoar's software components can be reused or inspected to serve as an inspiration, basis or benchmark for other soaring controller designs. We also present a live evaluation of ArduSoar showing its robust behavior in different environmental conditions.

\section{Soaring, Thermals, Thermal Estimation}

\noindent
Soaring usually refers to process of using a rising air mass to gain altitude by birds and aircraft. Technically, while in \emph{dynamic} soaring the source of lift is a wind gradient \cite{lawrance-icra09}, not rising air, here we focus on \emph{static} soaring, which always relies on air being pushed upwards in some way. 

\begin{figure}
\centering
\includegraphics[width=0.48\textwidth]{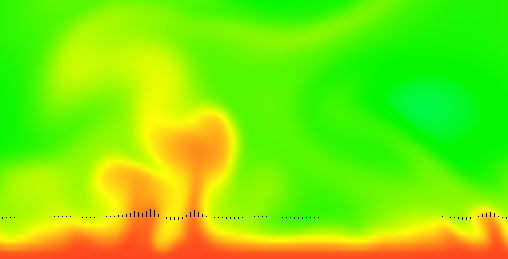}
\caption{\small Rayleigh-B\'enard convection simulation showing two rising thermal plumes (red color indicates hotter air). The lower boundary of the simulation heats the adjacent air hotter than the ambient air temperature. The hotter air gains buoyancy and rises in turbulent plumes to mix with the cooler air above.  The small black vertical bars indicate vertical air speed at an altitude and show the typical bell shaped distribution of uplift through the center of a thermal.}
\vspace{-0.4cm}
\label{fig:thermals_rb}
\end{figure}

The most commonly encountered type of rising air regions is a thermal. Thermals start immediately above parts of Earth's surface that accumulate more heat than surrounding areas, e.g. asphalt or plowed fields, and then emit the heat back into the atmosphere, warming up the air above them. They are invisible, but are frequently pictured as air columns, possibly slanted due to wind. Vertical air velocity in the center of such a column is higher, becomes lower towards the column fringes, and eventually turns into sink. In reality and even under realistic simulation models, thermals do not always conform to this idealized concept. Figure \ref{fig:thermals_rb} is a screenshot from a Rayleigh-B\'enard simulation (see, e.g., \citet{kadanoff-phys01}), a standard high-fidelity convection model that gives a more accurate idea of thermals' irregular and chaotic nature.

In all existing work on autonomous soaring in thermals, autopilots/soaring controllers use thermal models that can be estimated by observing the aircraft's vertical air velocity and combining this data with information about the aircraft's 3D position changes. Soaring controllers try to fit such a thermal model to this data and, if the fit is good enough, synthesize a trajectory that exploits the thermal's lift distribution, as predicted by the model, to gain altitude. The model at the heart of Rayleigh-B\'enard simulations has been used in this way to learn sailplane trajectories \cite{reddy-pnas16} for simulated aircraft, but is expensive computationally and nontrivial to tune.

Therefore, for computational and ease-of-use reasons, ArduSoar relies on the thermal model proposed by \citet{wharington-1998}. 
It makes a number of implicit assumptions, which we make explicit here, all of which apply to \emph{the thermal's horizontal cross-section at a given altitude}:

\begin{itemize}
\item{} A thermal is stationary: its position and distribution of lift within it, i.e., the air's vertical velocity, does not change with time. Note, however, that this assumption is for a fixed altitude only, and the model allows for a thermal's properties to vary with altitude due to air cooling down or due to the thermal getting affected by wind.

\item{} A thermal has a single ``center'', called \emph{core}, at position $x^{th}, y^{th}$.

\item{} The vertical air velocity within a thermal's horizontal cross-section has a bell-shaped distribution w.r.t. the thermal core. It is strongest at $(x^{th}, y^{th})$, and has speed $W^{th}$ m/s there. Thus, the lift strength at position $x, y$ is given by

\vspace{-0.3cm}
\begin{equation}
w(x,y) = W^{th} \exp \left( -\frac{(x-x^{th})^2 + (y - y^{th})^2}{{R^{th}}^2} \right),
\label{eq:thermal}
\end{equation}

where $R^{th}$ is a parameter describing the thermal radius.
\end{itemize}

Parameters $W^{th}$, $R^{th}$, $x^{th}$ and $y^{th}$ can be continually estimated using a Kalman filter 
if we assume a Gaussian distribution over them and keep updating this distribution with sensor observations. \citet{hazard-aiaa10} has used an Unscented Kalman filter (UKF) for this purpose. In ArduSoar, due to the computational constraints of the hardware platforms on which it typically runs, Pixhawk and APM, we use an Extended Kalman Filter (EKF) instead.

An EKF is a type of Kalman filter that allows maintaining a Gaussian estimate of the system state given a sequence of observations if the system transition function $f(\mathbf{X}, \mathbf{u})$ and observation function $h(\mathbf{X})$ are nonlinear. Specifically, suppose we start with a state distribution $\mathcal{N}(\mathbf{X}_k, P_k)$, and want to compute a new estimate $\mathcal{N}(\mathbf{X}_{k+1}, P_{k+1})$ after a system makes a transition over one time step and we receive an observation $\mathbf{o}$ about it. First 
 we linearize the model $f$ about the current state estimate by computing the Jacobian

\begin{equation}
F = \frac{\partial f}{\partial \mathbf{X}} \biggr\rvert_{\mathbf{X}_k} 
\label{eq:ekf_first}
\end{equation}

\noindent
Then we update the state estimate using the state transition

\begin{align}
\hat{\mathbf{X}}_{k+1} &= f(\mathbf{X}_k, \mathbf{u}_k) \\
\hat{P}_{k+1} &= F P_k F^\intercal + Q \label{eq:cov_upd}
\end{align}

\noindent
Then we linearize $h$ around this new state estimate by computing the Jacobian

\begin{equation}
H = \frac{\partial h}{\partial \mathbf{X}} \biggr\rvert_{\hat{\mathbf{X}}_{k+1}}
\label{eq:h_lin}
\end{equation}

\noindent
and calculate the approximate Kalman gain

\begin{equation}
K = \hat{P}_{k+1} H^\intercal (H \hat{P}_{k+1}H^\intercal +R)^{-1}
\label{eq:k_gain}
\end{equation}

\noindent
As the final step, the state distribution so far needs to be corrected for the observation process:

\begin{align}
\mathbf{X}_{k+1} &= \hat{\mathbf{X}}_{k+1} + K(\mathbf{o} - h(\hat{\mathbf{X}}_{k+1}))\\
P_{k+1} &= (I-KH)\hat{P}_{k+1}
\label{eq:ekf_last}
\end{align}

\noindent
Matrices $Q$ and $R$ are process and observation noise covariances, respectively, and are tunable parameters in our model.

\section{Algorithmic Aspects of ArduSoar}

Gaining altitude with a thermal's help can be broken down into four stages --- thermal detection, identification, exploitation, and exit. At an abstract level, an aircraft traveling through the air with its motor shut down will tend to lose altitude at a certain rate that depends on the aircraft's airspeed via a function called the \emph{polar curve} that we discuss shortly. The autopilot knows both the polar curve and the current airspeed, and therefore at any point can easily determine how fast the aircraft \emph{would be} descending if the air around it was still. If the aircraft is flying through a thermal, the air around it is not still; it is rising and lifting the aircraft with it. The autopilot can detect this by comparing the aircraft's actual altitude change rate to the one predicted by the polar curve, and enter the thermalling mode. In the thermalling mode, the autopilot continually keeps refitting the altitude change rate data it is receiving to a thermal model, e.g., \cref{eq:thermal} using an EKF, and trying to keep the aircraft within the thermal to gain altitude. If the altitude change rate becomes close to the ``natural'' sink rate predicted by the polar curve, it is an indication that the thermal is weak and it is time to exit it.

ArduSoar, our soaring controller integrated into an open-source fixed-wing \suav{} autopilot called ArduPlane \cite{arduplane-doc}, operationalizes this intuition using the following data streams from ArduPlane's Attitude and Heading Reference System (AHRS):

\begin{itemize}
\item{} $h$ - altitude
\item{} $\phi$ - roll (bank) angle
\item{} $v$ - airspeed
\item{} $x^{abs}$ - GPS latitude
\item{} $y^{abs}$ - GPS longitude
\item{} $v^{wind}$ - wind velocity
\end{itemize}

\noindent
We now describe ArduSoar's operation in each stage.

\subsection{Polar Curve and Variometer Data}

The key to ArduSoar's operation is a stream of \emph{variometer} data, synthesized from data streams described above. To a first approximation, a variometer aims to measure the vertical speed of the air around an aircraft, thereby helping the autopilot determine whether the aircraft is in a thermal. In practice, however, no onboard instrument has direct access to the true vertical speed of the air. A variometer tries to deduce it from the \emph{aircraft's own} vertical speed w.r.t. the ground, which, in the case of a sailplane with its motor turned off, can be imparted by the sailplane being in a thermal \emph{or} by pitching up or down.

Therefore, instead of measuring the vertical airspeed, we implement a variometer that measures the rate of change of a sailplane's \emph{total specific energy}. It is obtained by dividing the total energy of a sailplane in a glide (i.e. with the motor off, if present),

\begin{equation}
E = mgh + \frac{1}{2}mv^2,
\end{equation}

\noindent
by $mg$ and differentiating w.r.t. time. The result is 

\begin{equation}
\dot{e} = \dot{h} + \frac{v\dot{v}}{g}
\end{equation}

\noindent
Note that the measurement units of $\dot{e}$ are \si{m/s}. Computing variometer data involves estimating $\dot{h}$ and $\dot{v}$, which can be approximated by $\frac{\Delta e}{\Delta t}$ computed using successive $h$ and $v$ data points from the altitude and airspeed data streams.

To recover the true vertical speed of the air mass surrounding the sailplane, we need to further correct $\dot{e}$ with the aircraft's natural sink at its current airspeed $v$ and bank angle $\phi$. This corrected variometer is known as a \emph{netto variometer}, and has a value of $\SI{0}{m/s}$ in a steady glide through still air. The sink rate can be derived from the sailplane's aforementioned \emph{drag polar curve} and is given by

\begin{align}
v_z(v, \phi) = v \left( \frac{\mathbf{C_{D0}}}{C_{L}} + \frac{\mathbf{B}C_{L}}{\cos^2\phi} \right),
\label{eq:bcd0}
\end{align}

\noindent
where

\begin{equation}
C_L = \frac{\mathbf{K}}{v^2}.
\label{eq:liftcoeff}
\end{equation}

\noindent
In these expressions, $v$ is the current airspeed, $\phi$ the bank angle, and $\mathbf{C_{D0}}$, $\mathbf{K}$, and $\mathbf{B}$ are airframe-specific constants. Of these,
\begin{equation}
K = \frac{2mg}{\rho A_{\mbox{wing}}},
\label{eq:k}
\end{equation}
\noindent
where $m$ is the \suav{'s} mass, $A_{\mbox{wing}}$ is its wing area, and $\rho$ is the air density for the \suav{'s} operating conditions. $\mathbf{C_{D0}}$ and $\mathbf{B}$ need to be determined by fitting them to flight data using Equation \ref{eq:bcd0}.

The netto variometer signal can then be defined as:
\begin{equation}
\dot{e}_{net} = \dot{e} + v_z(v, \phi)
\end{equation}

\subsection{Thermal Detection}

ArduSoar constantly monitors a first order low-pass-filtered version of $\dot{e}_{net}$: 
\begin{equation}
\dot{e}_{filt_{n}} = \mathbf{T_c} \dot{e}_{net} + \left(1 - \mathbf{T_c} \right) \dot{e}_{filt_{n-1}}
\end{equation}
where $\mathbf{T_c}=0.03$. When $\dot{e}_{filt}$ exceeds a manually settable threshold (parameter $\mathtt{SOAR\_VSPEED}$ in ArduSoar's implementation), ArduSoar switches to the thermalling mode. Using this filtered version of the vario signal helps to remove turbulence which might otherwise trigger the algorithm.

\subsection{Thermal Identification and Exploitation}

ArduSoar naturally interleaves these two stages: it uses observation data --- variometer readings associated with GPS locations --- to update its distribution over possible thermal parameters, and at the same time chooses its trajectory --- a circle centered at a location $x_0, y_0$ of radius $\hat{R}^{th}$ --- based on the thermal parameter distribution.

Recall from Equation \ref{eq:thermal} that ArduSoar's thermal model has 4 parameters that need to be estimated: $W^{th}$, $R^{th}$, $x^{th}$ and $y^{th}$. We make the following simplifying assumption: \\ 

\noindent
\emph{Given wind velocity vector $\vec{v}^{wind}$, aircraft velocity vector $\vec{v}^a$, and  aircraft and thermal center locations $\vec{p}^a$ and $\vec{p}^{th}$ in the Earth reference frame, after time $dt$ their locations in the Earth reference frame are $\vec{p}^a = \vec{p}^{a'} + (\vec{v}^{wind} + \vec{v}^a) dt$ and $\vec{p}^{th'} = \vec{p}^{th} + \vec{v}^{wind} dt$ respectively.} \\

In other words, we assume that the aircraft and the thermal location are affected by wind in the same way. This isn't completely accurate \cite{telford-jas70}, because thermals tend to move slower than the surrounding air mass. Nonetheless, for small $dt$, i.e., high re-estimation frequency, it is a very good approximation of reality. Moreover, it allows us to easily carry out all calculations in the aircraft's reference frame, since, under this assumption, the relative displacement of the the aircraft and the thermal is entirely due to the aircraft's velocity vector. 

As in \citet{hazard-aiaa10}, we define the aircraft to be  at the origin of its own reference frame, which in turn lets us define 

\begin{equation}
    \mbox{Thermal state X} = \begin{bmatrix}
           W^{th} \\
           R^{th} \\
           x\\
           y
         \end{bmatrix} = \begin{bmatrix}
           \mbox{Strength} \\
           \mbox{Radius} \\
           \mbox{Center dist. north of \suav{}} \\
           \mbox{Center dist. east of \suav{}}
         \end{bmatrix}
  \end{equation}

Recall that state tracking with an EKF involves updating the current state distribution's mean with a linearization of the transition function $f$ (Equation \ref{eq:ekf_first}) and observation function $h$ (Equation \ref{eq:h_lin}). In our case, assuming that the aircraft has travelled distance $dx$ to the north and $dy$ to the east during a given time step,

\begin{equation}
\mathbf{X}_{k+1} = f(\mathbf{X}_{k}) = \mathbf{X}_{k} +  \begin{bmatrix}
           0 \\
           0 \\
           -dx \\
           -dy
         \end{bmatrix}.
\end{equation}

\noindent
Note that we have direct access only to the \suav{'s} GPS positions $x^{abs}, y^{abs}$ in the Earth reference frame. Since the \suav{} can be affected by wind, $dx \neq \Delta x^{abs}$ and $dy \neq \Delta y^{abs}$ for two successive $x^{abs}$ (resp. $y^{abs}$) measurements. Rather, 

\begin{align}
dx &= \Delta x^{abs} - v^{wind}_x \\
dy &= \Delta y^{abs} - v^{wind}_y,
\end{align}

\noindent
i.e., we need to perform \emph{wind correction} to obtain thermal motion w.r.t. the aircraft.

Since $f(\mathbf{X}_{k})$ effectively has no process dynamics, we get 

\begin{equation}
F = \frac{\partial f}{\partial \mathbf{X}} = \frac{\partial \mathbf{X}_{k+1}}{\partial \mathbf{X}} = \begin{bmatrix}1 & 0 & 0 & 0 \\
0 & 1 & 0 & 0 \\
0 & 0 & 1 & 0 \\
0 & 0 & 0 & 1
\end{bmatrix} = I
\label{eq:our_f}
\end{equation}

\noindent
For the observation function, the only readings relevant to the thermal state vector that we get is vertical air velocity $w$ at the aircraft's position w.r.t. the thermal center, so 

\begin{equation}
\mathbf{o} = w = h(\mathbf{X})= W^{th}exp\left(-\frac{x^2 + y^2}{{R^{th}}^2} \right)
\end{equation}

\noindent
and

\begin{equation}
H^\intercal = \frac{\partial h}{\partial \mathbf{X}}^\intercal= \begin{bmatrix}
           exp\left(-\frac{x^2 + y^2}{{R^{th}}^2} \right) \\
           \frac{2W^{th} (x^2 + y^2)}{{R^{th}}^3}exp\left(-\frac{x^2 + y^2}{{R^{th}}^2} \right) \\
           \frac{2W^{th}x}{{R^{th}}^2}exp\left(-\frac{x^2 + y^2}{{R^{th}}^2} \right)\\
           \frac{2W^{th}y}{{R^{th}}^2}exp\left(-\frac{x^2 + y^2}{{R^{th}}^2} \right)
         \end{bmatrix} 
         \label{eq:o}
\end{equation}

\noindent
Note two aspects that make ArduSoar's EKF update extra efficient:

\begin{itemize}
\item{} Due to Equation \ref{eq:our_f}, covariance update (Equation \ref{eq:cov_upd}) becomes simply $\hat{P}_{k+1} = P_k  + Q$.

\item{} Since our observation vector has only a single component, vertical air velocity, and $\frac{\partial h}{\partial X}$ is a $1 \times 4$ vector (Equation \ref{eq:o}), it follows that in Equation \ref{eq:k_gain} $(H \hat{P}_{k+1}H^\intercal +R)$ is a scalar value and the inversion during Kalman gain computation reduces to a single division.
\end{itemize}

ArduSoar's EKF update runs at 5Hz, giving ArduSoar a stream of thermal parameter estimates. In order to use this information to gain height, ArduSoar enters the autopilot into the LOITER mode (see the next section), in which ArduPlane ensures that the aircraft orbits a specified location in a circle of a specified radius. ArduSoar sets $(x,y)$ from the mean thermal estimate to be the center of this circle, but setting the radius $\hat{R}$ is less straightforward.

In theory, using the mean $R^{th}$ estimate, ArduSoar could choose $\hat{R}$ to strike the optimal balance between estimated lift $\hat{R}$ meters away from the thermal center and the lift loss due to bank angle required to stay in a turn of radius $\hat{R}$. In practice, however, we made $\hat{R}$ a manually settable parameter for ArduSoar, for reasons described in Section \ref{ssec:rad}.

\subsection{Thermal Exit}

ArduSoar constantly monitors the strength of the thermal to decide whether to exit and search for a better thermal. It is assumed that the sailplane is currently circling the thermal center at distance $\hat{R}$, and a lift estimate at this distance is calculated using Equation \ref{eq:thermal}. The estimate is then adjusted using a constant correction factor, $\mathbf{K_{sink}}$, representing sink from the drag polar and bank angle. Thus, ArduSoar computes $W^{th} exp\left(-\frac{x^2 + y^2}{\hat{R}^2}\right) - \mathbf{K_{sink}}$, and compares it against the same $\mathtt{SOAR\_VSPEED}$ threshold as in the thermal entry logic to determine if the thermal has become too weak to continue with. Note that this exit condition doesn't rely on the \emph{actual} lift measurements, only on those predicted by the thermal model. This approach has the advantage over other methods \cite{allen-2007,edwards-aiaa2008} of not being overly sensitive to the speed of the controller's response to fluctuations in thermal size and position.

\section{ArduPlane integration \label{sec:integr}}

Although a soaring controller can enable a \suav{} to stay aloft by extracting energy from the atmosphere, per se it lacks a lot of the useful functionality of a full-blown autopilot, including an AHRS system for estimating aircraft state, waypoint following capability, safety mechanisms such as geofencing, etc. Integrating ArduSoar into ArduPlane resulted in a flight controller for fixed-wing \suav{s} that has the rich feature set of a major open-source autopilot in addition to thermalling functionality. 

To integrate ArduSoar into ArduPlane, we have augmented ArduPlane with a number of parameters configurable via a ground control station. Some of them, such as $\mathtt{SOAR\_VSPEED}$, pertain to ArduSoar itself, while the rest control ArduSoar's interoperation with ArduPlane as shown in Figure \ref{fig:aplusa}. The full list of ArduSoar's parameters is given in Table \ref{t:params} in the Appendix.

In particular, ArduPlane has two modes relevant to ArduSoar, AUTO and LOITER. In AUTO mode, given a set of waypoints, a target airspeed, and several other parameters, ArduPlane's default behavior is to fly the aircraft fully autonomously, following the specified waypoint course. In LOITER mode, ArduPlane's default behavior is to orbit a point with specified GPS coordinates at a specified radius.

\noindent
We modified ArduPlane's AUTO mode to opportunistically use thermalling functionality as shown in Figure \ref{fig:aplusa}. If the aircraft just took off or if it descended to the altitude given by the new $\mathtt{SOAR\_ALT\_MIN}$ parameter (segment 1 of the flight in Figure \ref{fig:aplusa}), it climbs under motor to $\mathtt{SOAR\_ALT\_CUTOFF}$ altitude (segment 2 in Figure \ref{fig:aplusa}) while following pre-set waypoints. Upon reaching $\mathtt{SOAR\_ALT\_CUTOFF}$, it turns off the motor and glides down at ArduPlane's target airspeed (segment 3). During this stage, ArduSoar keeps monitoring the aircraft state. If it detects a thermal, it takes over navigational control and starts thermalling by putting the aircraft into LOITER mode (segment 4 in Figure \ref{fig:aplusa}). The point ArduSoar commands the aircraft to orbit is the current mean of the thermal center estimate, $(x^{th}, y^{th})$. The radius is given by a special ArduPlane parameter for LOITER mode called $\mathtt{WP\_LOITER\_RAD}$. While thermalling, the aircraft does not follow its waypoint course and may deviate from it. It may also reach an altitude given by the $\mathtt{SOAR\_ALT\_MAX}$ parameter. This safety parameter terminates thermalling to prevent the aircraft from climbing too high. At that point, ArduSoar stops thermalling, gives navigational control back to ArduPlane, and the aircraft starts another motorless glide (segment 5). This process can repeat many times without any human intervention.

\begin{figure}
\vspace{0.2cm}
\includegraphics[width=0.485\textwidth]{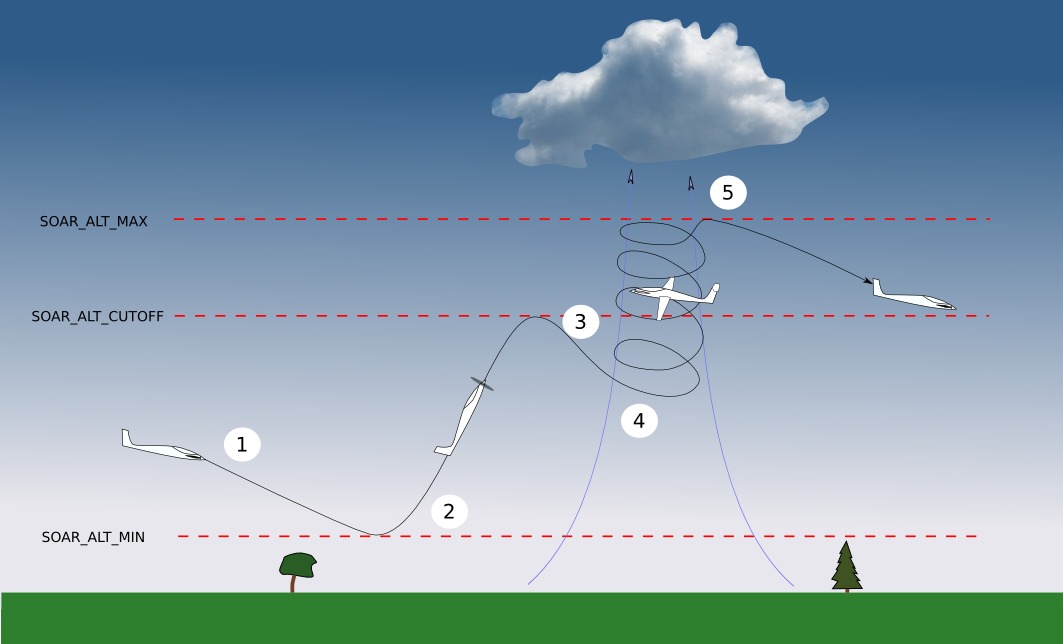}
\captionof{figure}{\small A fixed-wing \suav{'s} typical flight profile under ArduPlane's control with ArduSoar enabled. The \suav{} spends much of its flight gliding down with its motor off (flight segment \#1). If it descends to a preset minimum altitude specified by ArduPlane's $\mathtt{SOAR\_ALT\_MIN}$ parameter, it turns the motor on and climbs to $\mathtt{SOAR\_ALT\_CUTOFF}$ altitude (segment \#2). Then it turns the motor off and resumes the glide (segment \#3). If during the glide it detects a thermal, it activates ArduSoar to exploit it (segment 4). If it reaches a preset altitude limit $\mathtt{SOAR\_ALT\_MAX}$ in this way, it quits thermalling and starts gliding again (segment \#5).}
\label{fig:aplusa}
\end{figure}

\begin{figure}
\hspace{-0.2cm}
\includegraphics[width=0.505\textwidth]{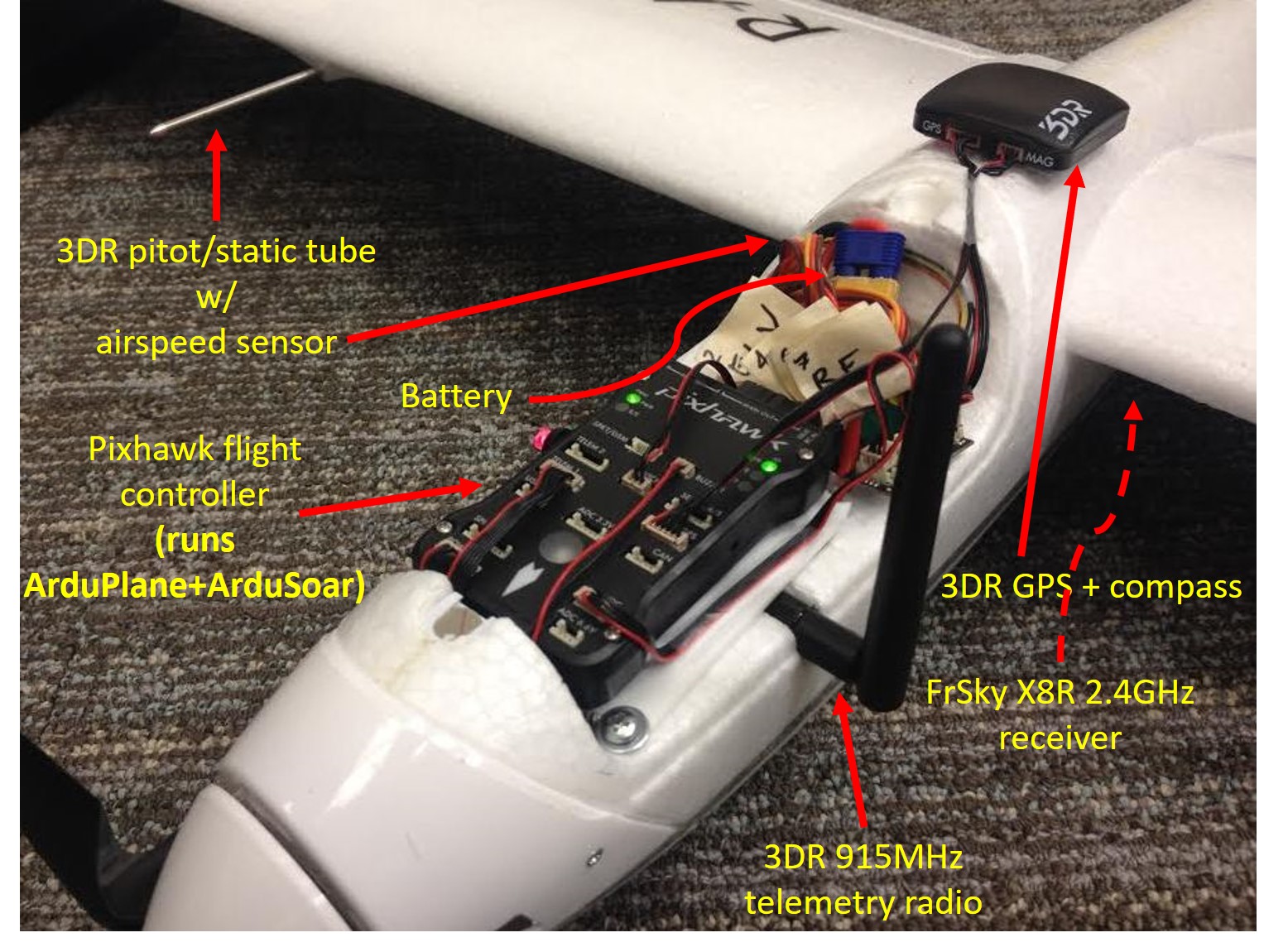}
\captionof{figure}{\small Radian Pro with the canopy removed to show the electronic equipment bay.}
\label{fig:eq_diag}
\vspace{-0.3cm}
\end{figure}

\begin{figure}
\vspace{-0.5cm}
\hspace{-0.5cm}
\includegraphics[width=0.55\textwidth]{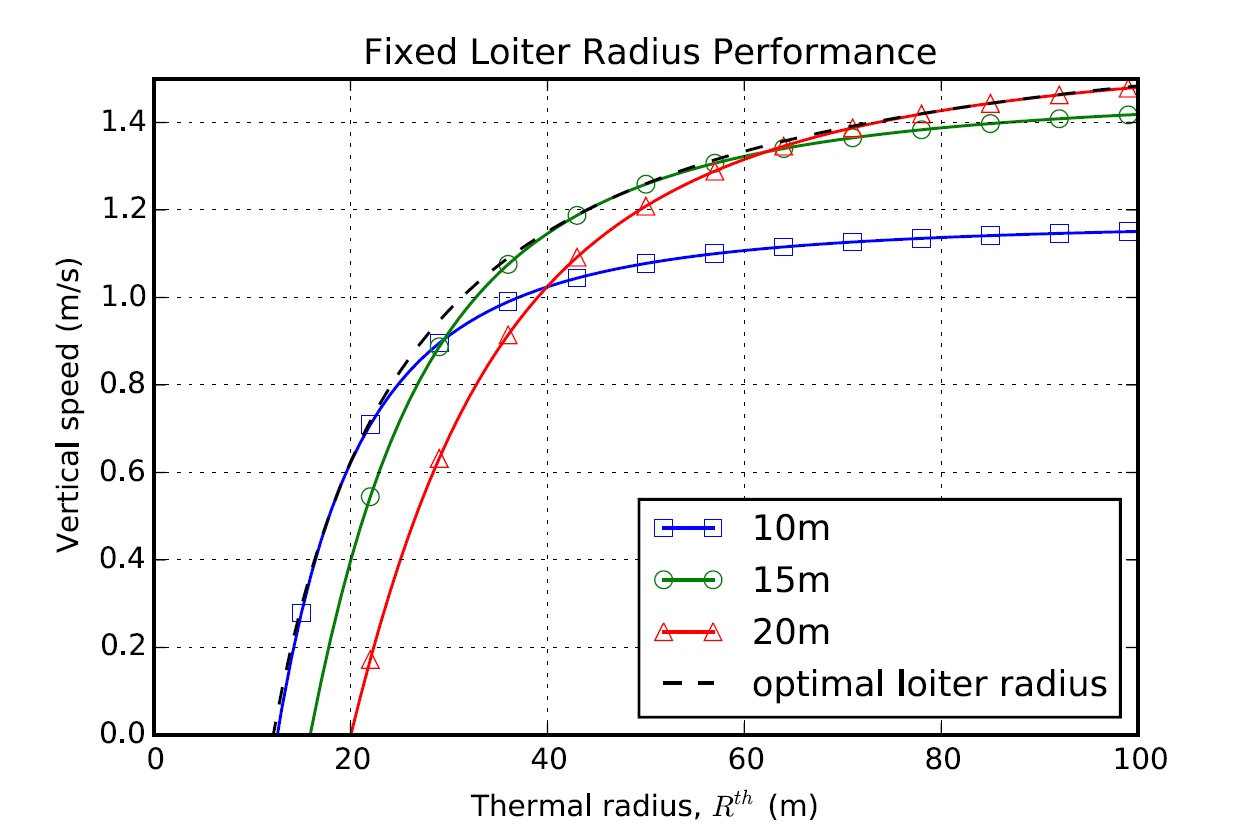}
\vspace{-0.6cm}
\caption{\small Soaring performance trade-off for 3 fixed loiter radii over a wide range of thermal sizes ($R^{th}$) with $W^{th}=2.5$ m/s. For each thermal radius $R^{th}$, the optimal loiter radius -- one that results in the highest altitude gain rate rate when taking into account the \suav{}'s bank angle -- is presented for comparison. Among fixed loiter radii, 15 m offers the best overall performance.}
\label{fig:loiterperf}
\vspace{-0.3cm}
\end{figure}

\section{Flight Testing \label{sec:tuning}}

The goal of flight testing was validation of ArduSoar's ability to gain altitude under a variety of real thermalling conditions. As a preliminary step, we estimated polar curve parameters of a Parkzone Radian Pro, a popular remote-controlled sailplane that served as our \suav{} testbed, and tuned parameters affecting ArduSoar's performance. The file with all these parameters, whose estimation is outlined below, is available at \emph{\href{https://github.com/Microsoft/Frigatebird}{https://github.com/Microsoft/Frigatebird}}.

\subsection{Equipment}

Our Radian Pro was equipped with a 3DR Pixhawk flight controller hardware (32-bit ARM processor, 256KB RAM, 2MB flash memory, 168MHz clock speed), a 3DR GPS and compass, a 3DR pitot/static tube and a sensor, a 3DR 915MHz telemetry radio, and a FrSky X8R receiver. The remote pilot could intervene at any point using a FrSky X9D+ 2.4GHz transmitter. The equipment diagram is shown in Figure \ref{fig:eq_diag}.

\subsection{Setting ArduSoar's Parameters}

The polar curve was estimated by conducting test glides in calm air. ArduPlane's AUTO mode was used to track a target airspeed around a triangular course, for a range of target airspeeds. The descent rate was measured for each, and ArduSoar parameters $\mathtt{SOAR\_POLAR\_B}$ and $\mathtt{SOAR\_POLAR\_CD0}$ representing constants $\mathbf{B}$ and $\mathbf{C_{D0}}$  in Equation \ref{eq:bcd0} were estimated from the resulting data using a least squares fit of that equation \cite{john1989introduction}. $\mathtt{SOAR\_POLAR\_K}$, corresponding to constant $K$ in Equations \ref{eq:liftcoeff} and  \ref{eq:k}, was determined directly from the wing measurements and weight of the Radian Pro, and $\rho=1.225$ kg/m$^3$ was assumed since all flight testing was conducted near sea level.

\begin{figure*}[tb]
\begin{subfigure}{0.5\textwidth}
\centering
\includegraphics[width=\textwidth]{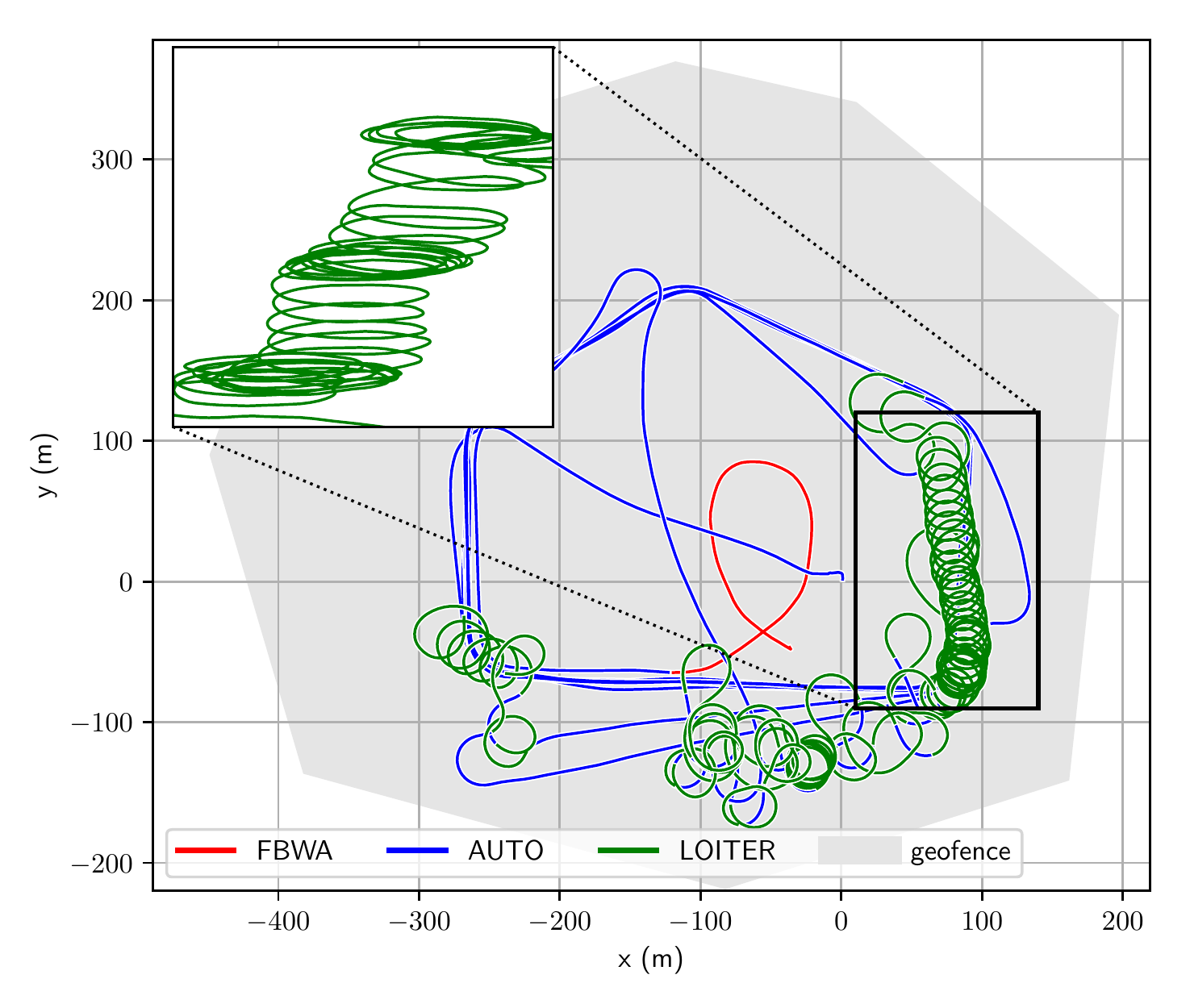}
\vspace{-0.7cm}
\caption{Steady thermals.}
\label{fig:windy_steady}
\end{subfigure}
\begin{subfigure}{0.5\textwidth}
\centering
\includegraphics[width=\textwidth]{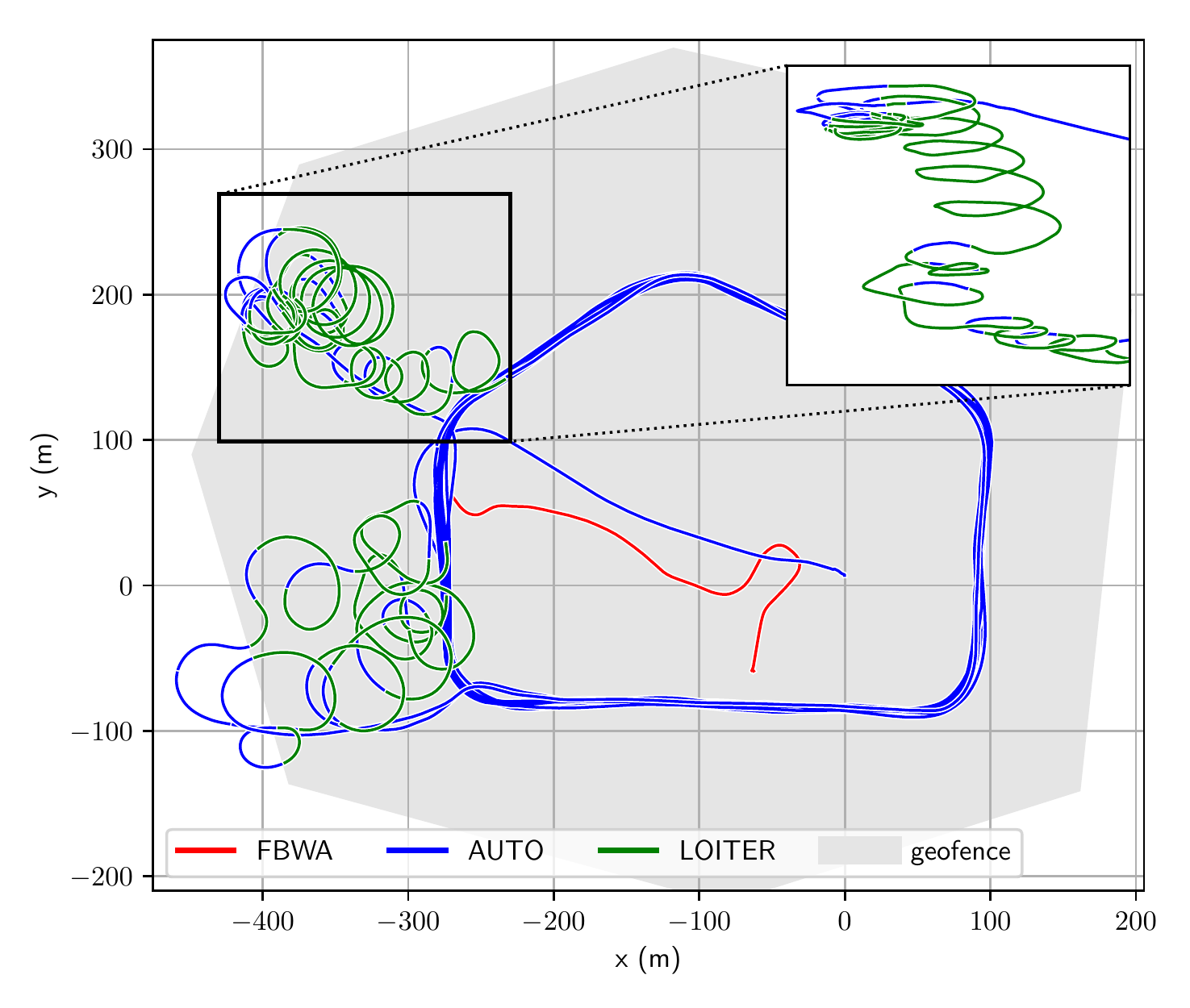}
\vspace{-0.7cm}
\caption{Windy, turbulent thermals.}
\label{fig:windy_turb_track}
\end{subfigure}
\begin{subfigure}{0.5\textwidth}
\centering
\includegraphics[width=\textwidth]{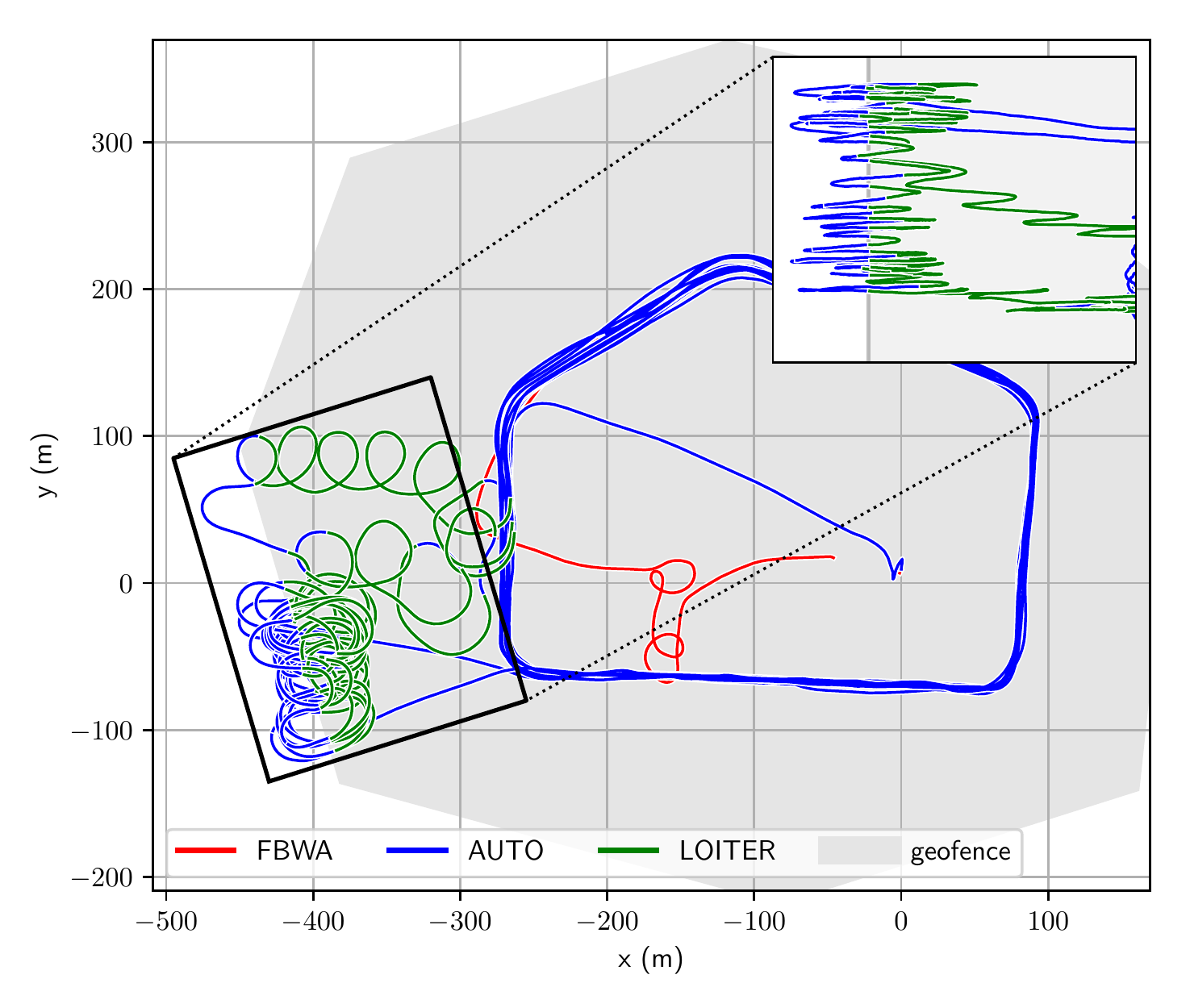}
\vspace{-0.7cm}
\caption{Windy orographic lift forcing the sailplane against the geofence.}
\label{fig:windy_oro_geofence_track}
\end{subfigure}
\begin{subfigure}{0.5\textwidth}
\centering
\hspace{1cm}
\includegraphics[width=0.82\textwidth,trim={2cm 1cm 0.5cm 1cm},clip]{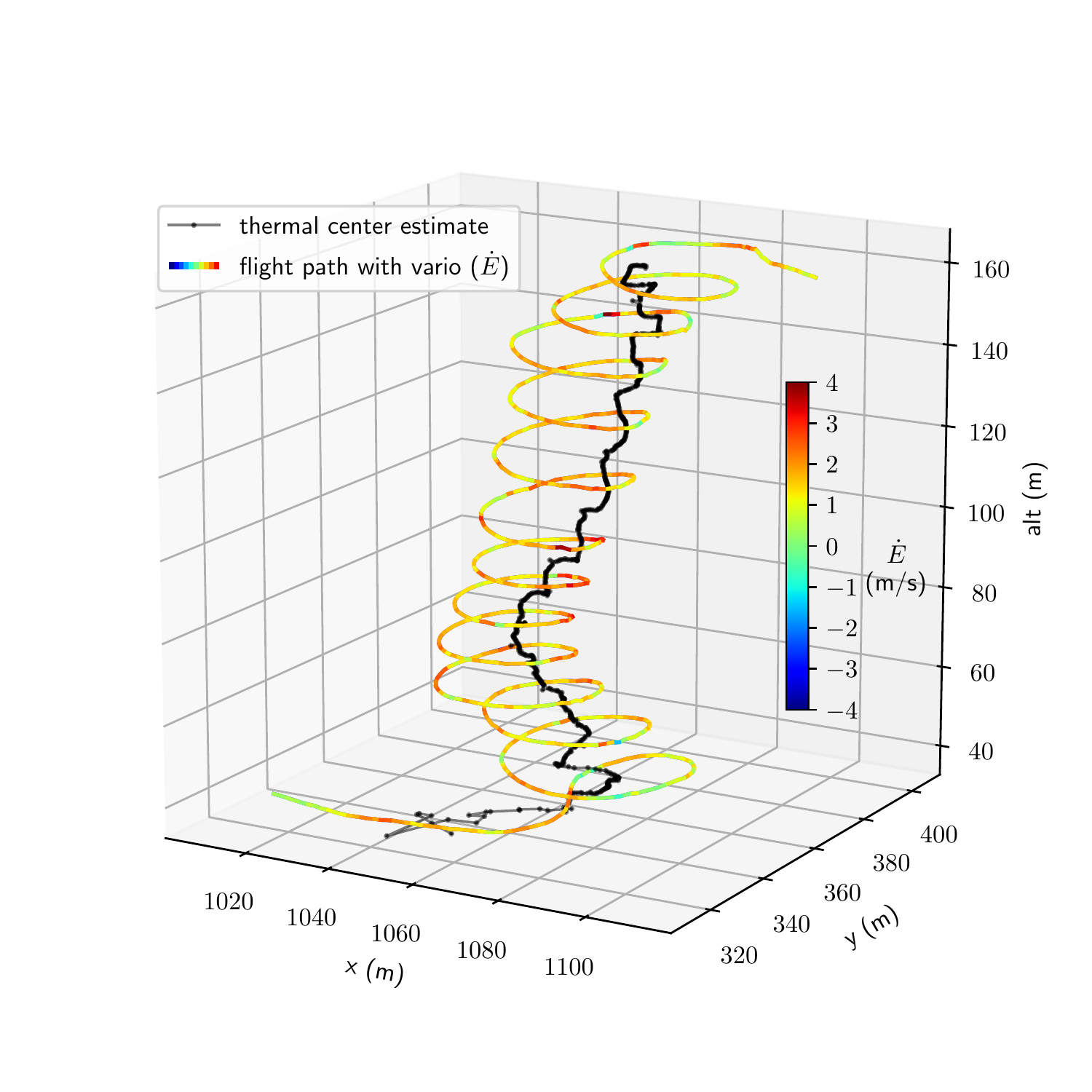}
\vspace{0.1cm}
\caption{\small Detail of a thermalling trajectory and EKF estimate of the thermal center, with the variometer signal indicated along the flight path.}
\label{fig:thermal_plot}
\end{subfigure}
\vspace{-0.1cm}
\caption{\small ArduSoar flight paths along fixed circuits in different soaring conditions, with green path sections indicating soaring mode, and a 3D detail of a real-flight thermalling trajectory. Inserts in the subfigures show side views of interesting trajectory segments.}
\label{fig:flight_cond}
\vspace{-0.5cm}
\end{figure*}

Controller tuning required deriving estimates of variometer measurement noise $R = [ \mathtt{SOAR\_R}^2]$, initial state values and variances $x_{\text{init}}$ and $P_{\text{init}}$ and state process noise $Q = \mathit{diag}(\mathtt{SOAR\_Q1}^2, \mathtt{SOAR\_Q2}^2, \mathtt{SOAR\_Q2}^2, \mathtt{SOAR\_Q2}^2)$, where $\mathtt{SOAR\_R}$, $\mathtt{SOAR\_Q1}$ and $\mathtt{SOAR\_Q2}$ are ArduSoar parameters. $\mathtt{SOAR\_R}$ captures variations in vario readings due to actual measurement noise, unmodelled aircraft dynamics and turbulence, and was therefore taken as the variance in readings during circling flight in turbulent air. 

The elements of $Q$ are selected to give good filter response characteristics in steady-state thermalling. The strength and radius estimates should change slowly (on the order of the period of thermal orbit) relative to the position estimates.

\begin{figure}
\hspace{-0.25cm}
\includegraphics[width=0.51\textwidth]{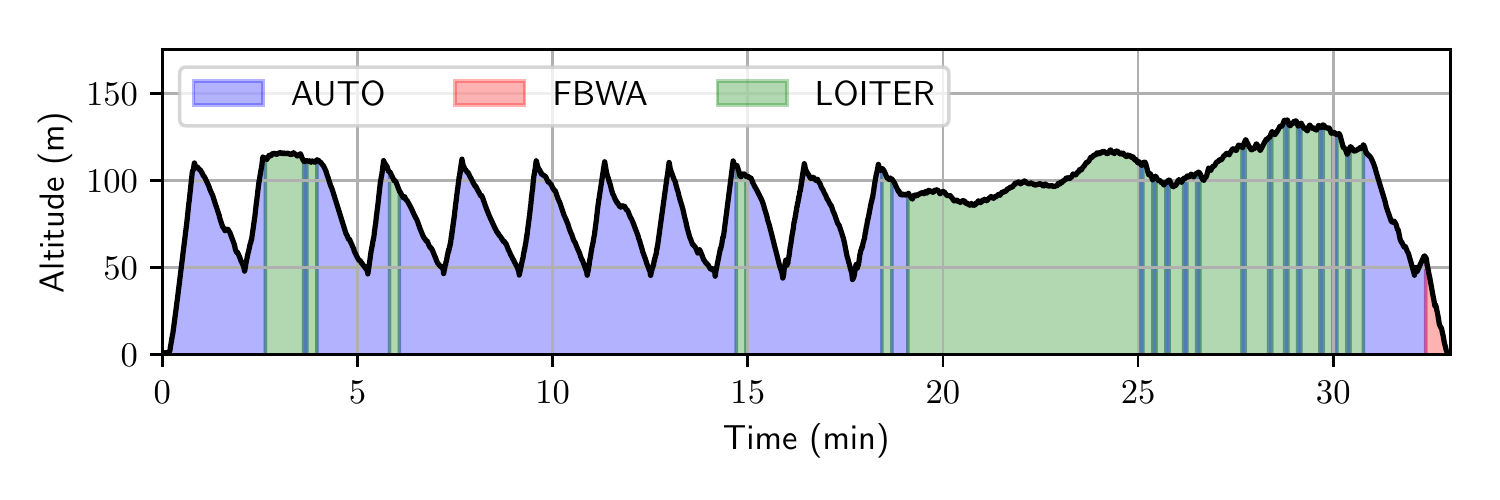}
\vspace{-0.7cm}
\caption{\small Plot showing altitude and flight mode vs. time for the flight in Figure \ref{fig:windy_steady}. The motor is on during climbs in AUTO mode, but is off for the entire duration of LOITER mode.}
\label{fig:messy_climb}
\vspace{-0.5cm}
\end{figure}

\subsection{Setting Thermalling Radius $\hat{R}$ \label{ssec:rad}}

Radius $R^{th}$ is an essential part of the model in Equation \ref{eq:thermal}, and its EKF-based estimates help determine the location of the thermal center. However, we have observed that after ArduSoar centers a thermal, the EKF develops growing  variances in $W^{th}$ and $R^{th}$ estimates and covariance between them, in effect becoming less and less confident whether the \suav{} is in a strong narrow thermal or a large weak one. This is not surprising given ArduSoar's lack of deliberate thermal exploration --- once it is confident of the center location, it orbits it and doesn't attempt to reduce the variance of distributions over $W^{th}$ and $R^{th}$. 

For this reason, rather than varying ArduSoar's thermalling radius $\hat{R}$ as a function of EKF's changing $R^{th}$ estimates, we set $\hat{R}$ to a fixed value specified by ArduPlane's $\mathtt{WP\_LOITER\_RAD}$ parameter. Note that in other strong soaring controllers \cite{allen-2007,andersson-jgd12}, thermalling radius is a manually set parameter as well. In our flight tests, for $\mathtt{WP\_LOITER\_RAD}$'s value we used the smallest turn radius that a Radian Pro can reliably maintain in LOITER mode, \SI{15}{m}. \Cref{fig:loiterperf}, based on thermal simulation, demonstrates the trade-off in soaring performance versus loiter radius and shows that this loiter radius value works well across a wide range of thermals. This is not surprising. Intuitively, using the smallest turn radius a given \suav{} supports in LOITER mode maximizes the \suav{'s} chances of exploiting small thermals while only insignificantly affecting its ability to exploit large ones. For large thermals, the \suav{} is likely to enter them away from the center, in a  relatively flat portion of the bell-shaped lift distribution (Equation \ref{eq:thermal}). In such a case, variometer readings collected from flying a small-radius circle may look too similar, making it difficult to determine where the thermal center is and potentially causing ArduSoar to loiter in a region of the thermal where lift is strong enough to stay in the thermal but that nonetheless is not near the actual thermal center. This results in the \suav{} gaining altitude at a lower vertical speed than the thermal allows. Nonetheless, even in these situations the \suav{} gains valuable potential energy.

\subsection{Qualitative Results}

With parameters set as above, we conducted dozens of flights to get a qualitative picture of the behavior of ArduSoar integrated into ArduPlane in various soaring conditions. Most of these flights were fully autonomous except landing: we set the Radian to follow fixed geofenced circuits along a set of waypoints in AUTO mode, from which the sailplane could deviate if it encountered thermals (see Figure \ref{fig:aplusa} and its caption in Section \ref{sec:integr}). In compliance with US FAA regulations, the flight altitudes were limited to $\approx$ \SI{120}{\meter} (400 ft) AGL in locations free of ground structures and to  $\approx$ \SI{120}{\meter} above structure height in the vicinity of structures. The flights took place from October 2017 to January 2018 in the US Pacific Northwest. When scheduling flights, we did not purposefully seek out days with favorable thermalling conditions. Indeed, all flights were carried out in at least 2 m/s wind and temperatures not exceeding 14\textdegree C, with at least partial cloud cover. At the same time, we avoided flying during rain. Overall, 21 flights resulting in catching at least one thermal.

Figures \ref{fig:windy_steady}-\ref{fig:windy_oro_geofence_track} show examples of resulting trajectories, with the AUTO-mode portion of the track, shown in blue, tracing out the main waypoint sequence, and the geofenced areas in grey. Figure \ref{fig:windy_steady} shows a flight in some of the best conditions we encountered --- steady thermals with some wind --- where ArduSoar mostly produced stable helix-like thermalling trajectories (green sections of the path). Conditions in Figure \ref{fig:windy_turb_track} were significantly more challenging due to thermals being irregular and turbulent, resulting in messy paths during thermalling that nonetheless yield altitude gains. Finally, conditions such as in Figure \ref{fig:windy_oro_geofence_track}'s pushed ArduSoar to the limits. Strong wind repeatedly pushed the sailplane against the geofence, which forced it out of the soaring mode (LOITER), but once back within the geofenced area the sailplane re-entered the soaring mode again and the process started again. In addition, the sailplane was likely in orographic or even wave lift --- note that it managed to enter the soaring in a large area along the bottom left part of the geofence. Thus, ArduSoar's thermal model was clearly violated in the flight in Figure \ref{fig:windy_oro_geofence_track}. In spite of this, ArduSoar still managed to gain altitude in LOITER mode, as the insert in that figure shows.

Figure \ref{fig:messy_climb} depicts an example altitude versus time plot for a flight with ArduSoar enabled. Although true parameters of real-world thermals are nearly impossible to obtain, precluding definitive statements about ArduSoar's absolute thermalling efficiency, this plot shows that ArduSoar can extend flight duration simply by using thermals to maintain altitude, as well as by gaining it if the thermal is strong enough.

\section{Related Work}

Besides soaring in thermals, an algorithm closely resembling ArduSoar has been used successfully for soaring in orographic lift \cite{fisher-bb16}, generated by wind blowing into a vertical obstacle such a tree line or a hill and forcing the air to move upwards. Although the true shape of a rising air mass in this case can be almost arbitrary and is certainly far from a column, ArduSoar's strategy of circling in order to stay within the air mass is \emph{a} valid approach to exploiting these updrafts.

The thermal model we use is due to Wharington et al \cite{wharington-1998}, who uses it in a simulation-based proof-of-concept reinforcement learning approach \cite{sutton-98}. The same model with minor modification was subsequently used in thermal soaring algorithms by \citet{allen-2007} and \citet{edwards-aiaa2008}, which differ in their approaches to model estimation. Like ArduSoar, both of them were flight-tested, but using a different hardware platform (Piccolo). While \citet{edwards-aiaa2008}'s controller is much too expensive for Pixhawk and ran on a laptop on the ground during these tests, transmitting soaring commands via a telemetry link, it managed to keep a sailplane in the air for 5.3 hours continuously. On the other hand, \citet{andersson-jgd12} provide a stability proof for a computationally lightweight controller based on Reichmann rules that can potentially be implemented on resource-constrained hardware such as Pixhawk. \citet{reddy-pnas16}, like \citet{wharington-1998}, present a proof-of-concept RL approach in simulation, but utilize the much more detailed thermal simulation based on Rayleigh-B\'enard convection. \citet{guilliard-rss18} compare ArduSoar to another thermalling controller, POMDSoar, that generates thermalling trajectories by approximately solving a POMDP and thereby explicitly plans for thermal exploration. In turbulent conditions of \citet{guilliard-rss18}'s experiments POMDSoar outperformed ArduSoar, but critically relied on accurate estimates of several additional airframe-specific parameters to do well, making POMDSoar more involved to use in practice.

An efficient alternative means of thermal identification for simulated autonomous thermalling is given by \citet{hazard-aiaa10}. This work derives a UKF-based thermal tracker and evaluates it assuming various open-loop environment exploration policies, such as flying a horizontal circular or sinusoidal pattern. ArduSoar uses Hazard's insights regarding thermal tracking using a KF variant. Another (extended) KF-based thermal tracker \cite{kahn-gcd17} uses a more sophisticated state dynamics model than ArduSoar, but is computationally more expensive as a result, and fails to capture uncertainty correlations between thermal position and strength estimates, reflected in the off-diagonal terms in the $P$ matrix of our estimator.
\section{Conclusions}

We have presented ArduSoar, an embedded thermal soaring controller that we implemented and tested in the codebase of a popular open-source autopilot for fixed-wing \suav{s} called ArduPlane. ArduSoar is based on a 4-variable EKF-based estimator with simple state dynamics to achieve a  computational cost low enough for ArduSoar to be executable on highly resource-constrained drone autopilot hardware such as Pixhawk or even APM. The implementation demonstrates good performance and has been used by several research groups and many individuals for leisure, research and commercial purposes.\\

\noindent
\textbf{Acknowledgements.} We would like to thank Rick Rogahn, Jim Piavis, Peter Braswell, and the ArduPilot developer community for their help in designing and testing ArduSoar's implementation.

\bibliographystyle{plainnat}
\bibliography{bibliography}
\appendix

\begin{table}[h]
\centering
\begin{tabular}{|p{3cm}|p{5cm}|}
\hline
Parameter name & Description \\ \hline 
$\mathtt{SOAR\_ENABLE}$ & Enables or disables ArduSoar functionality \\ \hline
$\mathtt{SOAR\_VSPEED}$ & A threshold for vertical air velocity that determines whether the aircraft that observes air moving upwards at $\mathtt{SOAR\_VSPEED}$ should enter/exit the thermalling mode. \\ \hline
$\mathtt{SOAR\_Q1}$ & Standard deviation of the process noise for thermal strength \\ \hline
$\mathtt{SOAR\_Q2}$ & Standard deviation of the process noise for thermal position and radius \\ \hline
$\mathtt{SOAR\_R}$ & Standard deviation of the observation noise \\ \hline
$\mathtt{SOAR\_DIST\_AHEAD}$ & Initial guess of the distance to the thermal center along the aircraft heading \\ \hline
$\mathtt{SOAR\_MIN\_THML\_S}$ & Minimum time the aircraft should keep trying to soar once it has entered the thermalling mode (LOITER), in seconds \\ \hline
$\mathtt{SOAR\_MIN\_CRSE\_S}$ & Minimum time the aircraft should avoid entering the thermalling mode (LOITER) after it last exited it, in seconds \\ \hline
$\mathtt{SOAR\_POLAR\_CD0}$ & Polar curve parameter (see Equation \ref{eq:bcd0}) \\ \hline
$\mathtt{SOAR\_POLAR\_B}$ & Polar curve parameter (see Equation \ref{eq:bcd0}) \\ \hline
$\mathtt{SOAR\_POLAR\_K}$ & Polar curve parameter (see Equation \ref{eq:liftcoeff})\\ \hline
$\mathtt{SOAR\_ALT\_MIN}$ & The minimum altitude above home position at which soaring or off-motor gliding is allowed, in meters. If the aircraft descends to $\mathtt{SOAR\_ALT\_MIN}$, it turns on the motor and starts climbing in AUTO mode.  \\ \hline
$\mathtt{SOAR\_ALT\_CUTOFF}$ & The altitude above home position, in meters, at which the aircraft turns off the motor after a climb from $\mathtt{SOAR\_ALT\_MIN}$ (or from the ground) and starts gliding in AUTO mode, trying to detect thermals. \\ \hline
$\mathtt{SOAR\_ALT\_MAX}$ & The maximum altitude above home position, in meters, that the aircraft is allowed to reach. Upon reaching it, the aircraft exits the thermalling mode if necessary and starts a glide in AUTO mode. \\ \hline
\end{tabular}
\caption{Parameters of the ArduSoar controller settable in ArduPlane.}
\label{t:params}
\end{table}
\end{document}